# Seamless Integration and Coordination of Cognitive Skills in Humanoid Robots: A Deep Learning Approach

Jungsik Hwang and Jun Tani

## Abstract

This study investigates how adequate coordination among the different cognitive processes of a humanoid robot can be developed through end-to-end learning of direct perception of visuomotor stream. We propose a deep dynamic neural network model built on a dynamic vision network, a motor generation network, and a higher-level network. The proposed model was designed to process and to integrate direct perception of dynamic visuomotor patterns in a hierarchical model characterized by different spatial and temporal constraints imposed on each level. We conducted synthetic robotic experiments in which a robot learned to read human's intention through observing the gestures and then to generate the corresponding goal-directed actions. Results verify that the proposed model is able to learn the tutored skills and to generalize them to novel situations. The model showed synergic coordination of perception, action and decision making, and it integrated and coordinated a set of cognitive skills including visual perception, intention reading, attention switching, working memory, action preparation and execution in a seamless manner. Analysis reveals that coherent internal representations emerged at each level of the hierarchy. Higher-level representation reflecting actional intention developed by means of continuous integration of the lower-level visuo-proprioceptive stream.

This work was supported by the National Research Foundation of Korea (NRF) grant funded by the Korea government (MSIP) [No. 2014R1A2A2A01005491]. Jungsik Hwang is with the school of Electrical Engineering in Korea Advanced Institute of Science and Technology, Daejeon, South Korea (e-mail: jungsik.hwang@gmail.com). Jun Tani is a corresponding author and he is a professor at the school of Electrical Engineering in Korea Advanced Institute of Science and Technology, Daejeon, South Korea and an adjunct professor at Okinawa Institute of Science and Technology, Okinawa, Japan. (e-mail: tani1216jp@gmail.com).

# I. INTRODUCTION

It would be desirable if a robot could learn to generate complex goal-directed behaviors from its own sensorimotor experience, as human beings do. One challenge in reaching this goal is that such complex behaviors require an agent to coordinate multiple cognitive processes. For instance, imagine a robot conducting an object manipulation task with a human partner, such as reaching for and grasping an object. The human partner indicates a target objects located on the workspace through a gesture. Then, the robot observes the workspace, finds the indicated object by combining the perceived gesture information as well as the perceived object's properties, switches its attention to the object, prepares an action and executes it. Even this simple task is complex, involving diverse cognitive skills such as visual perception, intention reading, working memory, decision-making, action preparation and execution. It is essential to link these skills with synergy by developing spatio-temporal coordination among them. Furthermore, these skills ideally arise from the robot's experience (of reaching for and grasping objects, for example), rather than from hand-engineered features reflecting a human engineer's understanding of what any given task may require [1-9].

In this study, we employed a deep learning approach to build a robotic system which can directly and autonomously learn from its own visuomotor experience. Deep learning is a fast-growing field in machine learning and artificial intelligence with remarkable advances, such as text recognition, speech recognition, image recognition and many others (See [10-12] for recent reviews of deep learning). One of the most important characteristics of deep learning is that deep networks can autonomously extract task-related features in high-dimensional data, such as images and action sequences, without the necessity of hand-engineered feature extraction methods [1, 10, 12]. So, deep learning provides an important tool for robotics, because through deep learning a robot can learn directly from its huge-dimensional sensorimotor data acquired through dynamic interaction with the environment [1]. A few recent studies [13-16] have demonstrated the plausibility of deep learning in the field of robotics. However, several challenges in adapting deep learning schemes to robotics remain. For example, a robotic system must process spatio-temporally dynamic patterns, whereas deep learning schemes have generally been designed to process static patterns. In addition, robotic tasks typically incorporate multiple sensory modalities, such as vision, proprioception and audition, while most deep learning applications attend to single modalities, e.g. visual facial recognition. Also, Sigaud and Droniou [1] have pointed out that it is still unclear how higher-level representations can be built by stacking several networks.

In this paper, we propose a dynamic deep neural network model called the Visuo-Motor Deep Dynamic Neural

Network (VMDNN) which can learn to generate goal-directed behaviors by coordinating multiple cognitive processes including visual perception, intention reading[1], attention switching, memorization and retrieval with working memory, action preparation and generation. The model was designed to process and to integrate dynamic visuomotor patterns directly perceived through the robot's interaction with its environment. The VMDNN is composed of three different types of subnetwork: the Multiple Spatio-Temporal scales Neural Network (MSTNN) [17], the Multiple Timescales Recurrent Neural Network (MTRNN) [18] and the PFC (Prefrontal Cortex) subnetworks. The MSTNN has demonstrated an ability to recognize dynamic visual scenes [17], and the MTRNN to learn compositional actions [18]. In the VMDNN model, these two subnetworks are tightly coupled via the PFC subnetwork, enabling the system to process dynamic visuomotor patterns simultaneously. In other words, the problems of perceiving visual patterns and generating motor patterns were regarded as inseparable problem and therefore the visual perception and the motor generation were performed simultaneously in a single VMDNN model in the current study. This approach is based on the previous studies which emphasized the importance of perception-action coupling in robotic manipulation [13] as well as in developmental robotics [40]. Each part of the architecture is imposed with different spatial and temporal constraints to enable the hierarchical computation of visuomotor processing in the lower-level and its abstraction in the higher-level. We conducted a set of synthetic robotics experiments to examine the proposed model and also to gain insight into mechanisms involved in learning goal-directed actions in biological systems. Here, it is worth noting that artificial neural networks are meant to model the essential features of the nervous system, not its detailed implementation [19-21].

In our experiments, a humanoid robot learned goal-directed actions from huge-dimensional visuomotor data acquired from repeated tutoring during which the robot's actions were guided by the experimenter. We investigated how a set of cognitive skills can be integrated and coordinated in a seamless manner to generate goal-directed sequential behaviors of the robot. Particularly, we focused on identifying the human gestures, reading intention underlying them, and generating corresponding sequential behaviors of the humanoid robot in a reaching-and-grasping task. In our experiment, the robot was trained to recognize human gestures and to grasp the target object indicated by the gestures. This task thus required a set of cognitive skills such as visual perception, intention reading, working memory, action preparation and execution. Recognizing gestures is more challenging than recognizing static images since both spatial and temporal information in the gesture need to be identified. Moreover, reading intention of others by observing their behavior has been considered as one of the core abilities required for social cognition [5, 22-24]. In addition, reaching and grasping are fundamental skills that have significant influences on the development of perceptual and cognitive abilities [5]. The reaching-and-grasping task has been extensively studied in child development [25, 26] as well as in robotics [19, 20, 27-29]. In a robotic context, they require robust perception and action systems as well as simultaneous coordination of a set of cognitive skills, making hand-designing features demanding and time-consuming [14]. Moreover, the visuomotor task in our experiment was not explicitly segmented into the different sub-tasks such as gesture classification and action generation by the experimenter. Therefore, the task requires the robot to adapt to different task phases autonomously by coordinating a set of cognitive skills in a seamless manner throughout the task. In addition, this task requires the robot to have working memory capability to keep the contextual information such that the robot could compare the categorized human intention with perceived object properties. This implies that simply mapping perception to action cannot perform the task successfully since it lacks contextual dynamics. We expected that the synergic coordination of the above mentioned cognitive skills would arise through end-to-end learning of the tightly coupled structure in which the multiple subnetworks densely interact.

---

[1] Here, the term of "reading" or "categorizing" is used rather than "recognizing" because recognition may involve reconstruction of patterns but categorization in the current study does not.

In the learning stage, the robot learned a task in the supervised end-to-end manner. In the testing stage, we examined the model's learning and generalization capabilities. Furthermore, the robot was examined under a visual occlusion experimental paradigm in which the visual input to the model was unexpectedly and completely occluded. This was to verify whether the proposed model was equipped with a sort of memory capability for maintaining task-related information. In addition, we analyzed the neuronal activation in order to clarify the internal dynamics and representations emerging in the model during different phases of operation.

The remaining part of the paper is organized as follows. In Section II, we review several previous studies employing deep learning schemes in the robotic context. In Section III, we introduce the proposed model in detail. Section IV and V are devoted to the experimental settings and the results respectively. Several key aspects and the implications of the proposed model are discussed in Section VI. Finally, we conclude our paper and indicate current and future research directions in Section VII.

## II. RELATED WORKS

Due to the remarkable success of deep learning in various fields, recent studies have attempted to employ deep learning in the field of robotics (See [1] for a recent review). For instance, Di Nuovo, et al. [15] employed a deep neural network architecture in order to study number cognition in a humanoid robot. The robot was trained to classify numbers by learning the association between an auditory signal and corresponding finger counting activity. They found that their model was quicker and more accurate when both modalities were associated. Similarly, Droniou, et al. [2] introduced a deep network architecture which could learn from different sensory modalities, including vision, audition and proprioception. During experiments, their model was trained to classify handwritten digits, and they demonstrated that learning across multiple modalities significantly improved classification performance. Yu and Lee [30] employed a deep learning approach on reading human intention. A supervised MTRNN model was employed in their experiments, and they showed that their model could successfully recognize human intention through observing a set of motions. Lenz, et al. [14] proposed a two-stage cascaded detection system to detect robotic grasps in an RGB-D view of a scene and conducted experiments on different robotic platforms. They found that their deep learning method performed significantly better than one with well-designed hand-engineered features. Pinto and Gupta [31] also addressed the problem of detecting robotic grasps, adopting a convolutional neural network (CNN) model to predict grasp location and angle.

Although these studies demonstrate the utility of deep learning in a robotic context, they most focused on robotic perception and robots were not directly controlled by deep learning schemes. A few recent studies have attempted to utilize deep learning to control a robot. Wahlström, et al. [16], addressed the pixels-to-torques problem by introducing a reinforcement learning algorithm that enabled their agent to learn control policy from pixel information. Deep auto-encoders and a multi-layer feedforward neural network were employed in their model, and a 1-link robotic pendulum was used as a testing platform. Their model was able to learn closed-loop policies in continuous state-action spaces directly from pixel information. Noda, et al. [32] introduced a deep auto-encoder-based computational framework designed to integrate sensorimotor data. A humanoid robot was used in their object manipulation experiments, and they showed that their model was able to form higher-level multimodal representations by learning sensorimotor information including joint angles, RGB images and audio data. Levine, et al. [13] proposed a deep neural network model which learned a control policy that linked raw image percepts to motor torques of the robot. They showed that the robot was able to conduct various object manipulation tasks by learning perception and control together in an end-to-end manner. Park and Tani [33] investigated how a robot could infer the underlying intention of human gestures and generate corresponding behaviors of a humanoid robot. In their work, an MTRNN model was employed and the robot could successfully achieve the task by extracting the compositional semantic rules latent in the various combinations of human gestures.

Several problems confront ongoing deep-learning research in robotics applications, as represented by limitations in existing studies such as relatively simple testing platform [16, 33], separate processing of individual modalities [32], and inability to handle temporal information [13]. In the current study, we aim to address these challenges with a deep dynamic neural network model for a humanoid robot which can process and integrate spatio-temporal dynamic visuomotor patterns in an end-to-end manner.

## III. THE DEEP NEURAL NETWORK MODEL

In this section, we describe the Visuo-Motor Deep Dynamic Neural Network (VMDNN) in detail. The proposed model was designed to process and to integrate direct perception of visuomotor patterns in a hierarchical structure characterized by different spatio-temporal constraints imposed on each part of the hierarchy. It has several distinctive characteristics. First, the model can perform low-level visuomotor processing without hand-engineered feature extraction methods by means of deep learning schemes. Second, the model processes dynamic visuomotor patterns in a hierarchical structure essential to cortical computation [15, 34]. Third, perception and action are tightly intertwined within the system, enabling the model to form multimodal representations across sensory modalities. The VMDNN model consists of three types of subnetworks: (1) MSTNN subnetworks for processing dynamic visual images, (2) MTRNN subnetworks for controlling the robot's action and attention and (3) a prefrontal cortex (PFC) subnetwork located on top of these two subnetworks which dynamically integrates them (Fig. 1).

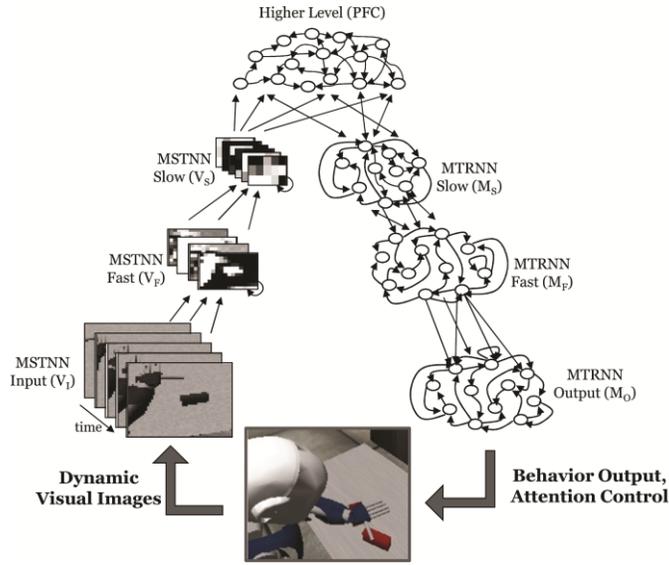

Fig. 1. The VMDNN model consists of three types of subnetworks: MSTNN subnetworks ($V_I$, $V_F$ and $V_S$) for dynamic visual image processing (left), MTRNN subnetworks ($M_O$, $M_F$ and $M_S$) for action generation (right) and the PFC subnetwork for integration of perception and action (top)

*A. MSTNN Subnetwork*

In our study, we employed Multiple Spatio-Temporal Scales Neural Network (MSTNN) to process dynamic visual images perceived by a robot conducting a visuomotor task. The MSTNN is an extended Convolutional Neural Network (CNN) [35] employing leaky integrator neural units with different time constants [17, 36]. Although conventional CNN models have been shown the ability to process spatial data such as static images, they lack the ability to process spatio-temporally dynamic patterns. To successfully conduct a visuomotor task, the robot needs to extract both spatial and temporal features latent in the sequential observations. Unlike conventional CNN models that utilize spatial constraints only, the MSTNN model has been shown that it can process both spatial and temporal patterns by imposing multiple spatio-temporal scales constraints on local neural activity [17, 36]. Consequently, the MSTNN can extract the task-related visual features latent in the dynamic visual images while the robot was tutored for the task iteratively.

The MSTNN subnetwork was composed of three layers imposed with different spatio-temporal constraints: the MSTNN-input ($V_I$) layer containing a current visual scene, the MSTNN-fast ($V_F$) layer with shorter-distance connectivity and smaller time constants, and the MSTNN-slow ($V_S$) layer with longer-distant connectivity and larger time constants. Each MSTNN layer organizes into a specific set of feature maps retaining spatial information of visual input. They are connected successively from $V_I$ to $V_S$.

## B. MTRNN Subnetwork

In the current study, we employed Multiple Timescales Recurrent Neural Network (MTRNN) for generating the robot's behavior and controlling the attention of the robot. The MTRNN is a hierarchical neural network model consisting of a multiple continuous time recurrent neural networks with leaky integrator neurons [18]. MTRNN has been shown superior performance in modeling robot's sequential action by utilizing its temporal hierarchy. To be more specific, the lower level in the MTRNN has a smaller time constant showing fast dynamics whereas the higher level has a bigger time constant exhibiting slow dynamics. Due to this temporal hierarchy, an MTRNN can learn compositional action sequences as a meaningful functional hierarchy emerges within the system [18, 34, 37]. Consequently, the entire behavior of the robot including reaching and grasping as well as visual attention control can be decomposed into a set of primitives for their flexible recombination adapting to various situations [18].

In our model, the MTRNN subnetwork is composed of three layers characterized by different temporal constraints: the MTRNN-slow ($M_S$) showing slow dynamics with the larger time constant, the MTRNN-fast ($M_F$) showing fast dynamics with smaller and the MTRNN-output ($M_O$) layer with the smallest time constant. Neurons in the $M_S$ and $M_F$ layers are asymmetrically connected to each other and to themselves. The $M_O$ layer is composed of groups of softmax neurons indicating the sparse representation of the model's output. The $M_O$ layer receives inputs from the $M_F$ layer and generates behavior outputs as well as attention control signals.

## C. PFC Subnetwork

At the top, the PFC (Prefrontal Cortex) layer tightly couples the MSTNN and MTRNN subnetworks for achieving tight association between the visual perception and the motor generation. The PFC layer is a recurrent neural network consisting of a set of leaky-integrator neurons equipped with recurrent loops in order to process abstract sequential information. The PFC layer receives inputs from the $V_S$ layer in the MSTNN subnetwork as well as from the $M_S$ layer in the MTRNN subnetwork, meaning that both abstracted visual information ($V_S$) and proprioceptive information ($M_S$) are integrated in the PFC layer. The PFC layer also has a forward connection to the $M_S$ layer to control the robot's behavior and attention.

The PFC layer can be characterized by several key features. First, neurons in the PFC layer are assigned the largest time constant. As a result, the PFC subnetwork exhibits the slowest-scale dynamics and this enables the PFC subnetwork to carry more information about a situation [38]. Second, neurons in the PFC layer are equipped with recurrent connections which are essential to handle dynamic sequential data [1, 9, 39]. Third, perception and action are coupled via the PFC layer which integrates two monomodal subnetworks (MSTNN and MTRNN) and builds higher-level multimodal representations from abstracted visuomotor information in each pathway. Lungarella and Metta [40] argued that perception and action are not separated but tightly coupled, with this coupling is gradually getting refined during developmental process.

*D. Problem Formulation*

Fig. 1 illustrates the structure of the VMDNN model. The input to the model ($I^t$) is an observation of the world at time step *t* which can be obtained from the robot's camera at the beginning to the end of the task. The observation is a pixel-level image represented as a matrix $H \times W$, where *H* is a height and *W* is a width of the image. Robot's behavior outputs as well as attention control signals are generated at the output layer $M_O$ at each time step *t*. Let $y^t = [y_1^t, y_2^t, \ldots, y_n^t]$ denotes the output of the model at the time step *t* where *n* is the number of neurons at the output layer $M_O$.

In forward dynamics computation, the problem is defined as to the behavior output and the attention signal ($y^t$) at time step *t* given a visual observation $I^t$ and the model parameters $\theta$ such as kernels, weights, and biases. In the training phase, the problem is defined as to optimize the model's parameters $\theta$ in order to minimize the error *E* at the output layer $M_O$ represented by the Kullback-Leibler divergence between the teaching signal $\bar{y}^t$ and the network's output $y^t$. The training data was visuomotor sequences obtained from the repeated tutoring prior to the training phase. The detailed description about the forward dynamics and the training phase are described in the following sections.

*E. Forward Dynamics for Action Generation*

The internal states of all neural units were initialized with neutral values at the onset of the action generation mode. Then, a pixel image $I^t$ in grayscale obtained from the robot's camera was given to the vision input layer ($V_I$) and neural unit's internal states and activations were successively computed in every layer from the input layer ($V_I$) to the output layer ($M_O$). The neuron activation at the $M_O$ layer was transformed to analog values using the softmax activation to control the robot's joints and attention. A detailed description of the computational procedure follows.

At each time step t, the internal state $u_i^{txy}$ and the dynamic activation $v_i^{txy}$ of the neural unit located at the (x, y) position in the *i*th feature map in the MSTNN layers (i ∈ $V_F \vee V_S$) are computed according to the following formulas:

$$u_i^{txy} = \left(1 - \frac{1}{\tau_i}\right)u_i^{(t-1)xy} + \frac{1}{\tau_i}\left[\sum_{j\in V_j}(k_{ij} * v_j^t)_{xy} + b_i\right] \quad (1)$$

$$v_i^{txy} = 1.7159 \times \tanh\left(\frac{2}{3}u_i^{txy}\right) \quad (2)$$

$\tau$ is the time constant, $V_j$ is the feature maps in the previous layer (if i ∈ $V_F$, then $V_j$ = $V_I$ and if i ∈ $V_S$, then $V_j$ = $V_F$), * is the convolution operator, $k_{ij}$ is the kernel connecting *j*th feature map in $V_j$ with the *i*th feature map in the current layer, and b is the bias. Please note that the hyperbolic tangent recommended in [41] was used as an activation function to enhance convergence.

From the PFC layer to the $M_O$ layer, the internal state $u_i^t$ and the dynamic activation $y_i^t$ of the *i*th neuron in the PFC and MTRNN layers (i ∈ PFC ∨ $M_S$ ∨ $M_F$ ∨ $M_O$) are determined by the following equations:

$$u_i^t = (1 - \frac{1}{\tau_i})u_i^{t-1} + \begin{cases} \frac{1}{\tau_i}\left[\sum_{j\in V_S} k_{ij} * v_j^t + \sum_{k\in M_S \vee PFC} w_{ik}y_k^{t-1} + b_i\right] & if\ i \in PFC \\ \frac{1}{\tau_i}\left[\sum_{j\in PFC} w_{ij}y_j^t + \sum_{k\in M_F \vee M_S} w_{ik}y_k^{t-1} + b_i\right] & if\ i \in M_S \\ \frac{1}{\tau_i}\left[\sum_{j\in M_S \vee M_F} w_{ij}y_j^{t-1} + b_i\right] & if\ i \in M_F \\ \frac{1}{\tau_i}\left[\sum_{j\in M_F} w_{ij}y_j^t + b_i\right] & if\ i \in M_O \end{cases} \quad (3)$$

$$y_i^t = \begin{cases} 1.7159 \times \tanh\left(\frac{2}{3}u_i^t\right) & if\ i \in PFC \vee M_S \vee M_F \\ \frac{\exp(u_i^t)}{\sum_{j\in M_O}\exp(u_j^t)} & if\ i \in M_O \end{cases} \quad (4)$$

$w_{ij}$ are the weight from the *j*th neural unit to the *i*th neural unit.

In sum, the image obtained from the robot's camera was sent to the input layer ($V_I$) of the model at each time step of the action generation mode. Then, the internal states and activations were successively computed from the input layer ($V_I$) to the output layer ($M_O$). The robot was operated based on the output of the model including the joint position values. After the execution of an action, the image acquired from the robot's camera was sent to the input layer ($V_I$) of the model again in the next time step. In this sense, an image given to the model can be considered as a visual feedback since it reflects the effect of the robot's action in the preceding time step.

*F. Training Phase*

The model was trained in a supervised end-to-end learning fashion. The training data consisted of raw visuomotor sequences obtained from repeated tutoring during which the robot was manually operated by the experimenter. From the beginning to the end of tutoring, a visual image perceived from the robot's camera (visual observation) was jointly collected with the encoder values of the robot's joint positions as well as the level of grasping and the level of foveation at each time step. The model was trained to abstract and associate visual perception with proprioceptive information using these visuomotor sequence patterns. Backpropagation through time (BPTT) [42] was employed in learning values of parameters, such as the kernels, weights and biases of the model.

Prior to end-to-end training, learnable parameters in the visual pathway (from $V_I$ to PFC) were initialized by means of pre-training. Studies have demonstrated that pre-training is an efficient method for initializing network parameters [10, 13, 43]. The pre-training method in our study is similar to that of [13] in which the visual part of the model was pre-trained prior to the end-to-end learning phase of the experiment. During pre-training in our study, the softmax output layer was connected to the PFC layer and the connections between the MTRNN subnetworks and the PFC layer, as well as the recurrent connections within the PFC layer were removed. In this condition, the system operates as an MSTNN model, and it was trained as a typical classifier using BPTT as described in [36]. Then, the values of the parameters in the visual pathway acquired during pre-training were used as initial values for the parameters in the same pathways during the end-to-end training phases of the experiment.

After the pre-training, end-to-end training was conducted from the $V_I$ layer to the $M_O$ layer. During end-to-end training, the model's entire learnable parameters (k, w and b) were updated to minimize the error E at the $M_O$ layer represented by Kullback-Leibler divergence between the teaching signal $\bar{y}_i^t$ and the network's output $y_i^t$.

$$E = \sum_t \sum_{i \in M_O} \bar{y}_i^t \log \frac{\bar{y}_i^t}{y_i^t} \qquad (5)$$

A stochastic gradient descent method was applied during end-to-end training and the entire learnable parameters were updated when each training data was presented as follows.

$$w_{ij}(n+1) = w_{ij}(n) - \eta \left( \frac{\partial E}{\partial w_{ij}} + 0.0005 w_{ij}(n) \right) \qquad (6)$$

$$k_{ij}(n+1) = k_{ij}(n) - \eta \left( \frac{\partial E}{\partial k_{ij}} + 0.0005 k_{ij}(n) \right) \qquad (7)$$

$$b_i(n+1) = b_i(n) - \eta \left( \frac{\partial E}{\partial b_i} \right) \qquad (8)$$

n is an index of the learning step and η is the learning rate. The weight decay method was used to prevent overfitting

[35] with the weight decay rate of 0.0005.

IV. EXPERIMENT SETTINGS

*A. Robotic Platform*

iCub [44] is a child-like humanoid robot consisting of 53 degrees of freedom (DOFs) distributed in the body. We used a simulation of the iCub [45] in our experiments. The iCub simulator accurately models the actual robot's physical interaction with the environment, making it an adequate research platform for studying developmental robotics [15, 45]. Simulation of the iCub is shown in (Fig. 2). In our simulation environment, a screen was located in front of the robot to display human gestures, the task table was placed between the screen and the robot and the two objects were placed on the task table.

At each time step, the interfacing program captured the image perceived from the robot's camera, preprocessed the captured image, sent it to the VMDNN model, received the output of the VMDNN model and then operated the robot based on the output of the model. To be more specific, the pixel image obtained from the robot's camera embedded in its left eye was passed to the vision input layer ($V_I$) of the model at each time step. The obtained image was preprocessed by resizing to 64 (w) × 48 (h), converting to grayscale, and normalizing to -1 to 1. The softmax values of the output layer ($M_O$) were converted into the values of the corresponding joint positions (7 DOFs in the right arm, 2 DOFs in the neck) as well as into grasping and attention control signals. Then, the interfacing program operated the robot based on the VMDNN model's outputs using the motor controller provided in the iCub software package [45]. Then, the interfacing program captured an image from the robot's camera and sent it back to the VMDNN model in the next time step.

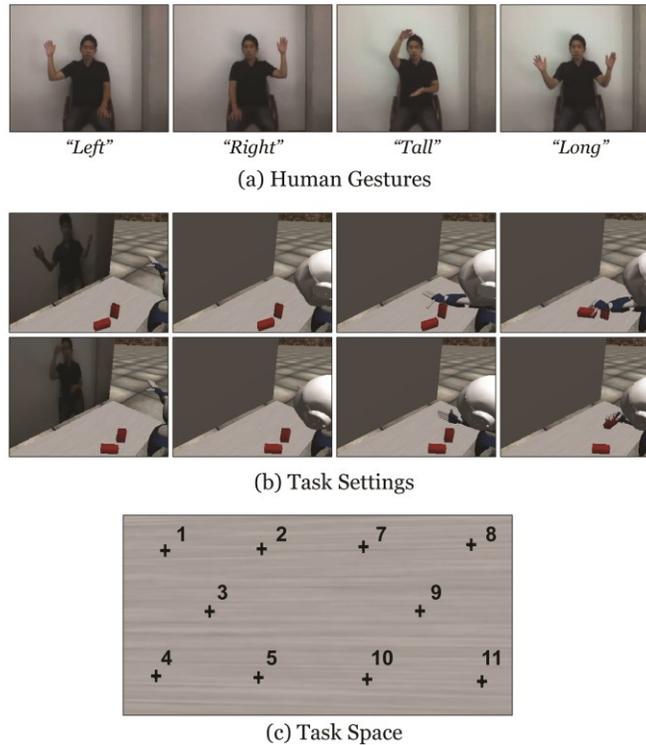

Fig. 2. The experimental settings using an iCub simulator. (a) Four types of human gestures (b) Two objects consisting of a tall and a long object were placed in the task space, and the robot grasped the target object indicated by a human gesture (c) Object locations on a task space.

Regarding the output for the robot's behavior, we used the robot's right arm consisting of 7 DOFs (shoulder's pitch, roll, yaw, elbow, wrist's pronosupination, pitch, and yaw). In addition, the network output the level of extension or flexion of finger joints in order to control grasping similar to [20]. The level varied from 1 (not grasping) to 10 (fully grasping) and it was used to control the 9 DOFs of the right hand. In our experiments, the VMDNN model also controlled visual attention which is essential to generate adequate robot's behavior [37]. Jeong, et al. [37] demonstrated that MTRNN could seamlessly coordinate visual attention and motor behaviors. In their work, the MTRNN model outputted the category of the object to be attended to, and the external visual guiding system localized the position of the specified object based on the network's output and the retina image so that the robot's camera could fixate the object. In contrast, no external module was employed in our study, and the VMDNN model itself controlled robot's visual attention. More specifically, two attention control mechanisms were employed in our study: attention shift and foveation. First, the robot located the object of attention at the center of the visual scene by orienting its head (2 DOFs – neck's pitch and yaw). Second, the robot controlled the resolution of the visual scene given to the network by controlling the level of foveation from 1 (minimum foveation) to 10 (maximum foveation). For instance, the level of foveation increased while the robot's hand was closely approaching to the attended object. Consequently, images containing the target object and the robot's hand were given to the network with higher resolution so that the model could more clearly perceive the object's properties including orientation and the location of the hand.

*B. Network Configuration*

The VMDNN model was composed of 7 layers: the $V_I$, $V_F$, $V_S$ layers in the MSTNN subnetwork, the PFC layer, and the $M_S$, $M_F$, $M_O$ layers in the MTRNN subnetwork. The structure of the VMDNN model used in this study was found empirically in our preliminary experiments [46]. Note that the structure of the VMDNN model including the number of layers in each subnetwork can be extended depending on the complexity of the task since the 'deeper' structure can enhance learning of complex functions in visuomotor patterns [11]. For instance, more number of layers in the MSTNN subnetwork can be employed to process more complex visual images [17]. Similarly, more complex robot's behavior can be learned by employing more number of layers in the MTRNN subnetwork as reported in [33].

Each MSTNN layer consisted of a set of feature maps retaining the spatial information of the visual input. The number and size of feature maps varied between layers. The vision input layer ($V_I$) had a single feature map containing the current visual scene (64 (w) × 48 (h)) obtained from the left eye of the robot. The $V_F$ layer consisted of 4 feature maps with sized 15 × 11 and the $V_S$ layer consisted of 8 feature maps with sized 5 × 3. The size of kernels in the $V_F$ and $V_S$ layers were set to 8 × 8 and 7 × 7 respectively. The sampling factors denoting the amount of shift of the kernel in a convolution operation were set to 4 and 2 for the $V_F$ and $V_S$ respectively. The PFC layer was composed of 20 neurons and the kernel size was 5 × 3 with the sampling factor set to 1. The numbers of neurons employed in the MTRNN layers ($M_S$, $M_F$ and $M_O$) were 30, 50 and 110 respectively. The 110 neurons in the $M_O$ layer were comprised of 11 groups of softmax neurons representing the 11 categories of the model's output: the joint position values of the robot's neck (2), the joint position values of the right arm (7), the level of grasping (1) and the level of foveation (1). The softmax output values in each group at the $M_O$ layer were inversely transformed to analog values that directly set the joint angles of the robots, the level of grasping and the level of foveation.

Regarding the time scale properties, we compared two different types of visual pathway: CNN and MSTNN. In the CNN condition, the time constants of both the $V_F$ and the $V_S$ layers were set to 1, resulting in no temporal hierarchy. On the other hand, in the MSTNN condition, the time constants of the $V_F$ and the $V_S$ layers were set to 1 and 15 respectively, resulting in a temporal hierarchy in the visual pathway. We also compared two different temporal scales (fast and slow) in the PFC layer. The time constant was set to 1 in the fast PFC condition whereas it was set to 150 in the slow PFC condition. In sum, there were 4 different network conditions examined in our experiments. Throughout the experiments, the time constants of the $M_S$, $M_F$ and $M_O$ layers were fixed to 70, 2 and 1 respectively. The proper values for the time constant at each level of the model were found heuristically in our preliminary study [46].

Prior to end-to-end learning, learnable parameters were initialized with the values acquired from the pre-training stage to enhance learning capability. During the pre-training stage, the model was trained to grasp an object without human gestures. Also, the visual pathway was additionally pre-trained to classify the four types of human gestures as described in Section III. During the end-to-end learning, the network was trained for 13,000 epochs with a learning rate of 0.01.

*C. Task*

The objective of the task was to grasp the target object indicated by the human gesture displayed on the screen at the beginning of the task. The overall task flow was as follows. At the beginning of the task, the robot was set to the home position, orienting its head toward the screen in front of the robot and stretching both arms sideways. From the home position, the robot observed a human gesture being displayed on the screen (Fig. 2). After observing the gesture, the robot oriented its head to the task table on which two objects consisting of a long and a tall object were placed. While observing the task space, the robot figured out the target object indicated by the human gesture and oriented its head to the target object. Then, the robot reached out its right arm and grasped the target object. For example, when the human gesture indicated the tall object, the robot had to pick out the tall object from among the two available along with information about its orientation. Similarly, when the human gesture indicated the right side, the robot had to figure out the type and orientation of the object on the right side (see the supplementary video). This task, therefore, inherently required the robot to have working memory capability to maintain the human gesture information throughout the task phases and dynamically combine it with the perceived object properties.

The visuomotor sequence for each training trial consisted of images perceived from the robot's camera (visual observation), the values of robot's joint positions and the values of the grasping and foveation signals and they were collected simultaneously from the beginning to the end of tutoring. Consequently, the images in the visuomotor sequences included the ones that were perceived while observing the human gestures on the screen as well as the ones perceived while completing the behavioral task. Similarly, the values of the robot's joint positions in the visuomotor sequences included the ones that were recorded while observing the human gestures as well as the ones recorded while acting in the task space after the observation. The robot was trained with 200 trials consisting of varying object configurations and human gestures. Two objects consisting of one tall and one long object were presented in each trial and the location and the orientation of the two objects were controlled so that the way of presenting the two objects did not bias the robot's behavior toward certain reaching and grasping behaviors and away from others. Regarding human gestures, a target object was specified by one of four different human gestures indicating the location (either left or right) or the type (either tall or long) of the target object (Fig. 2 (a)). A 40-frame video clip of the human gesture was displayed on the screen in front of the robot. We collected several gesture trials from 7 human subjects and pseudo-randomly selected from amongst them so that each type of gesture appeared the same number of times in the training dataset. Throughout the experiments, two different types of box-shape objects were used: a tall object with a size of 2.8cm × 5cm × 10cm and a long object with a size of 2.8cm × 10cm × 5cm. Objects were placed with 5 different orientations (-45°, -22.5°, 0°, 22.5°, 45°) at 10 positions symmetrically distributed on the XY-plane of the task space (Fig. 2 (c)).

During the testing stage, we evaluated the model's performance with respect to the learned trials (TR) as well as with a set of novel situations in order to examine the model's generalization capability. First, we tested the model with 60 trials in which two objects were randomly located (OBJ) to examine whether the robot was able to generalize reaching and grasping skills to unlearned object positions and orientations. We also examined the model's generalization capability with respect to human gestures, using 200 training trials with the gestures of a novel subject (SUB). Then, we examined the model's generalization capability with 60 cases of the most novel situation, in which randomly located objects were indicated by gestures of the novel subject (OBJ × SUB). It should be noted that main focus was on the model's ability to coordinate cognitive skills such as gesture recognition and working memory rather than solely on gesture recognition, since the latter is more concerned with typical perceptual classification tasks. In addition, we evaluated the model under a visual occlusion experimental paradigm in which the vision input to the network was completely occluded. The main focus of this evaluation was to verify whether the network was equipped with a sort of internal memory, enabling it to show robust behavior even the visual information was unexpectedly and completely occluded. From the training trials (TR) and the testing trials (OBJ × SUB), we selected 80 and 60 trials respectively. During the visual occlusion experiment, vision input to the network was occluded at the onset of observing the task space (t = 51), at the onset of attending to the target object (t = 55), at the onset of observing the target object (t = 61), at the onset of reaching (t = 66), and during reaching (t = 86). Prior to the testing, the learnable parameters were initialized to the ones obtained from the training.

## V. RESULTS

### A. Generalization Performances

Table 1 shows the success rate of each network condition. Each trial was evaluated as "successful" if the robot grasped and lifted an object and "failure" otherwise. In general, the MSTNN vision with the slow PFC condition showed the better performance than did other conditions. In this condition, the model successfully learned the training trials (TR = 98.5%) and was able to generalize learned skills to different testing conditions. This model was able to generalize reaching and grasping skills when the objects were randomly located (OBJ = 85.0%), when the gestures of the novel subject were displayed (SUB = 96.5%), and when the randomly located objects were specified by the novel subject (OBJ × SUB = 83.3%). It is worth noting that the model demonstrated relatively low success rates when in the fast PFC conditions, especially CNN vision with the fast PFC condition. The task required the robot to maintain human gesture information displayed at the beginning of the trial throughout all task phases. In the end, the model equipped with the temporal hierarchy performed significantly better than did those without it.

TABLE 1
THE SUCCESS RATE OF FOUR NETWORK CONDITIONS

| Network Conditions | | Testing Conditions | | | |
|---|---|---|---|---|---|
| *Type of Vision Layer* | *PFC Scales* | *TR* | *OBJ* | *SUB* | *OBJ × SUB* |
| *CNN* | *Fast* | 47.0% | 41.7% | 44.0% | 41.7% |
| *CNN* | *Slow* | 87.5% | 85.0% | 87.5% | 80.0% |
| *MSTNN* | *Fast* | 92.0% | 60.0% | 86.0% | 56.7% |
| *MSTNN* | *Slow* | 98.5% | 85.0% | 96.5% | 83.3% |

TABLE 2
THE PERCENTAGE OF TASK FAILURE CAUSED BY CONFUSION

| Network Conditions | | Testing Conditions | |
|---|---|---|---|
| Type of Vision Layer | PFC Scales | TR | OBJ × SUB |
| CNN | Fast | 59.4% | 71.4% |
| CNN | Slow | 44.0% | 41.7% |
| MSTNN | Fast | 68.8% | 65.4% |
| MSTNN | Slow | 0.0% | 20.0% |

We analyzed failure cases in training trials (TR) and in the most novel trials (OBJ × SUB). Particularly, we focused on task failure caused by apparent confusion between gesture-indicated and other objects. In these cases, the robot simply grasped the incorrect object. Table 2 shows the percentage of the error caused by confusion in the four network conditions. The model showed the least confusion errors in the MSTNN with slow PFC condition. Moreover, the error caused by confusion was more pronounced in fast PFC than in slow PFC conditions. This result implies that the model was able to maintain correct and stable representations by means of slow dynamics in the PFC layer.

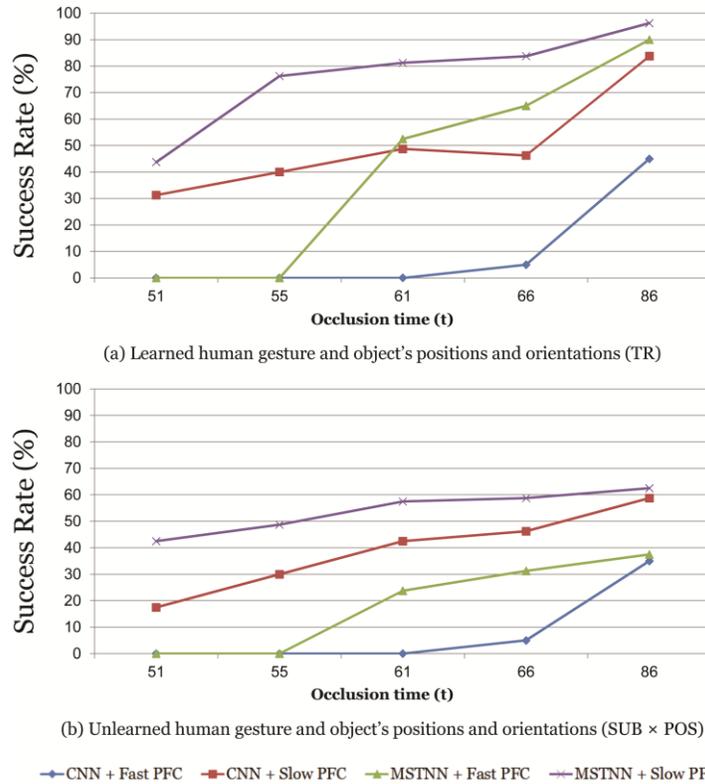

Fig. 3. The success rate of the four different network conditions in the visual occlusion experiment: (a) testing on learned human gestures and object positions and orientations (TR) and (b) testing on unlearned human gestures and object positions and orientations (SUB × POS).

Fig. 3 illustrates the success rate for each network condition with respect to (a) training trials (TR) and (b) testing trials (OBJ × SUB) with different occlusion timings. When vision input to the model was occluded, performance of the four different network conditions differed significantly. As expected, the model's performance in all conditions generally degraded when vision input was occluded during earlier task phases. Especially, the model showed relatively worse performance in the CNN vision with the fast PFC condition than other conditions. Therefore, it can be inferred that memory capability achieved by the slow dynamics subnetwork at the higher level plays an important role when vision input is occluded. That is, by means of memory, the robot was able to maintain information about human gestures as well as information about the target objects including position, type and orientation throughout the all phases of task execution. This result shows the importance of the internal contextual dynamics of the proposed model and highlights a difference between the proposed model and the previous study [13] which was prone to occlusion due to the lack of capability of keeping memory.

*B. Development of Internal Representation*

In order to reveal the model's self-organized internal representation, we analyzed neural activation during training trials using the t-Distributed Stochastic Neighbor Embedding (t-SNE) dimensionality reduction algorithm [47]. During t-SNE analysis, the parameters for the initial dimensions and the perplexity were set to 10 and 30 respectively. Fig. 4 depicts the internal representation emerging at each layer during the three different task phases. Each point indicates a single training trial and distances between those points represent relative similarity between trials. The color and the shape of each point denote the type of human gesture and the type of the object respectively. The number next to each point indicates the object's position.

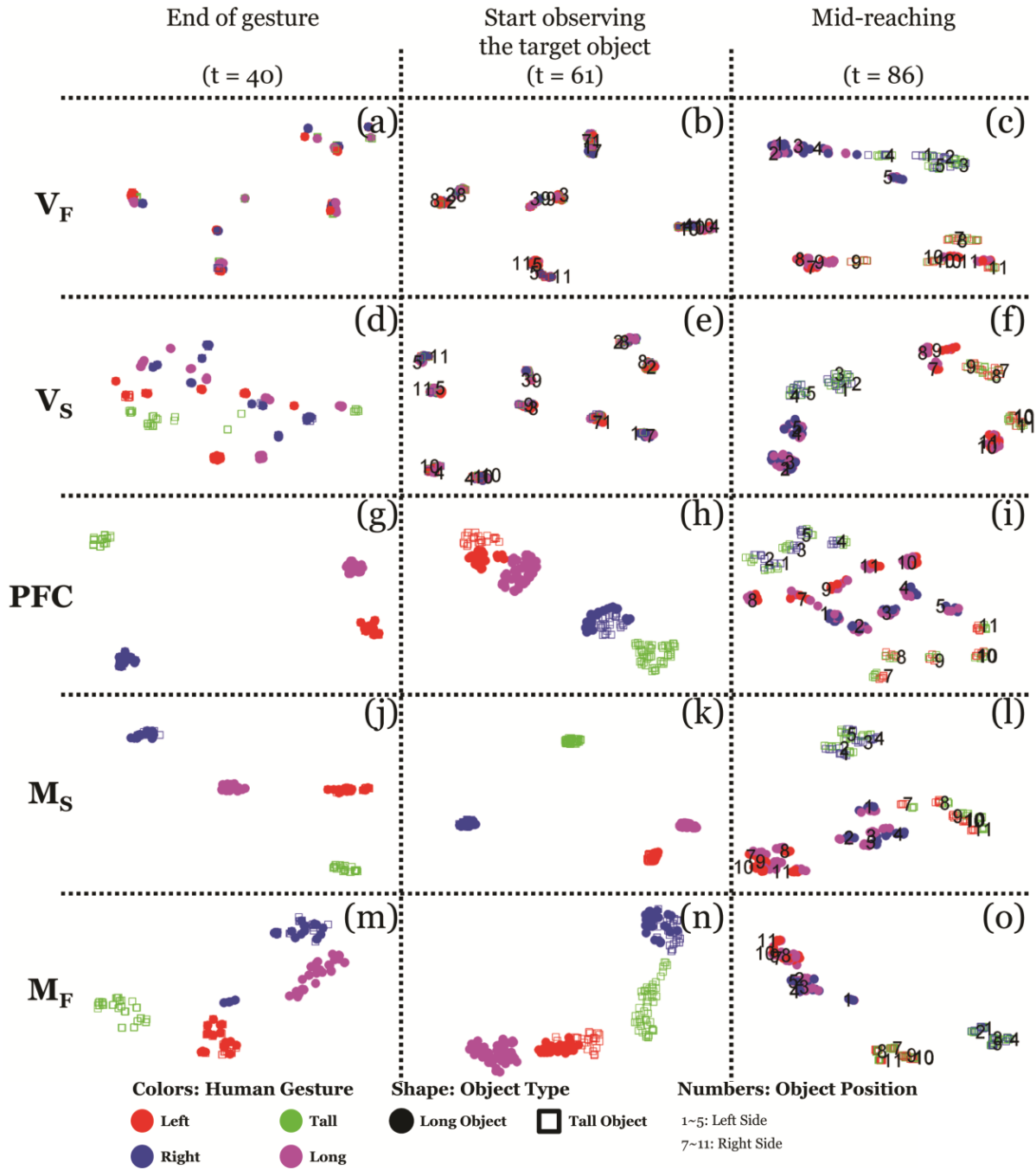

Fig. 4. Internal representations emerging during three different task phases at each layer in the MSTNN with the slow PFC. Each point represents a single trial and the distance between points represents the similarity between trials. The shape and the color of each point indicate the type of the target object and the type of human gesture, respectively. The number next to each point signifies the object's position. In some cases, numbers are omitted due to overlap. Please note that the mode of grasping is different depending on the type of object. We focused on the relationships between patterns, so we did not plot the axes varying between plots.

In MSTNN layers ($V_F$ and $V_S$), sequential visual images were abstracted in both temporal and spatial dimensions through hierarchical processing. For instance, where the robot begins observation of the task space, five clusters can be observed in the $V_F$ layer (b). These clusters correspond to the five possible pair locations employed during training.

Objects appearing in the same position appeared in the same cluster regardless of the type of object and type of human gesture. In the $V_S$ layer, those representations were further separated, with the type of target object differentiated within the clusters reflecting the relative locations of the two objects (e). Similarly, in the mid-reaching phase, the $V_F$ layer (c) and the $V_S$ layer (f) encoded both the type and the location of the target object, but the distinction with respect to the object's type was less clear in the $V_F$ layer.

Transitions of representations from those reflecting types of human gesture to those reflecting specific target objects organized in the PFC layer. After observing the gesture (g), four clusters reflecting the type of gesture appeared, suggesting that human gestures were successfully recognized. Then, those internal representations started to develop progressively, and smaller clusters indicating specific target objects emerged in the mid-reaching phase regardless of the presented human gesture (i). Our interpretation is that the higher-level of the model read the human's intention and translated it into robot's own intention to reach and to grasp the specified object. To be more specific, lower-level visual images containing human gestures were processed hierarchically by MSTNN subnetworks and the abstracted representation underlying one of the four human's intentions appeared in the higher-level (the PFC layer). Then, the robot simultaneously incorporated the perceived object's properties including location, type and orientation, and it formed its own intention for reaching and grasping the target object. In the end, this result suggests that the robot's intention was not explicitly mapped but arose dynamically from perceived information.

In the $M_S$ layer, four clusters representing each type of gesture appeared at the end of human gesture observation (j) and at the onset of target object observation (k). This implies that higher-level proprioception was calibrated based on the perceived human gestures before the robot exhibited different behaviors depending on the target object. At mid-reach (l), the $M_S$ layer encoded both the type and the location of the target object, but differences with respect to the object's location were less clear than those appearing in the PFC layer. In the $M_F$ layer, the internal representations developed similarly, but less clearly than those in the $M_S$ layer. For instance, the $M_F$ layer encoded both the type and the location of the object during mid-reach (o), but representations were less clear than those of the PFC (i) and $M_S$ (l).

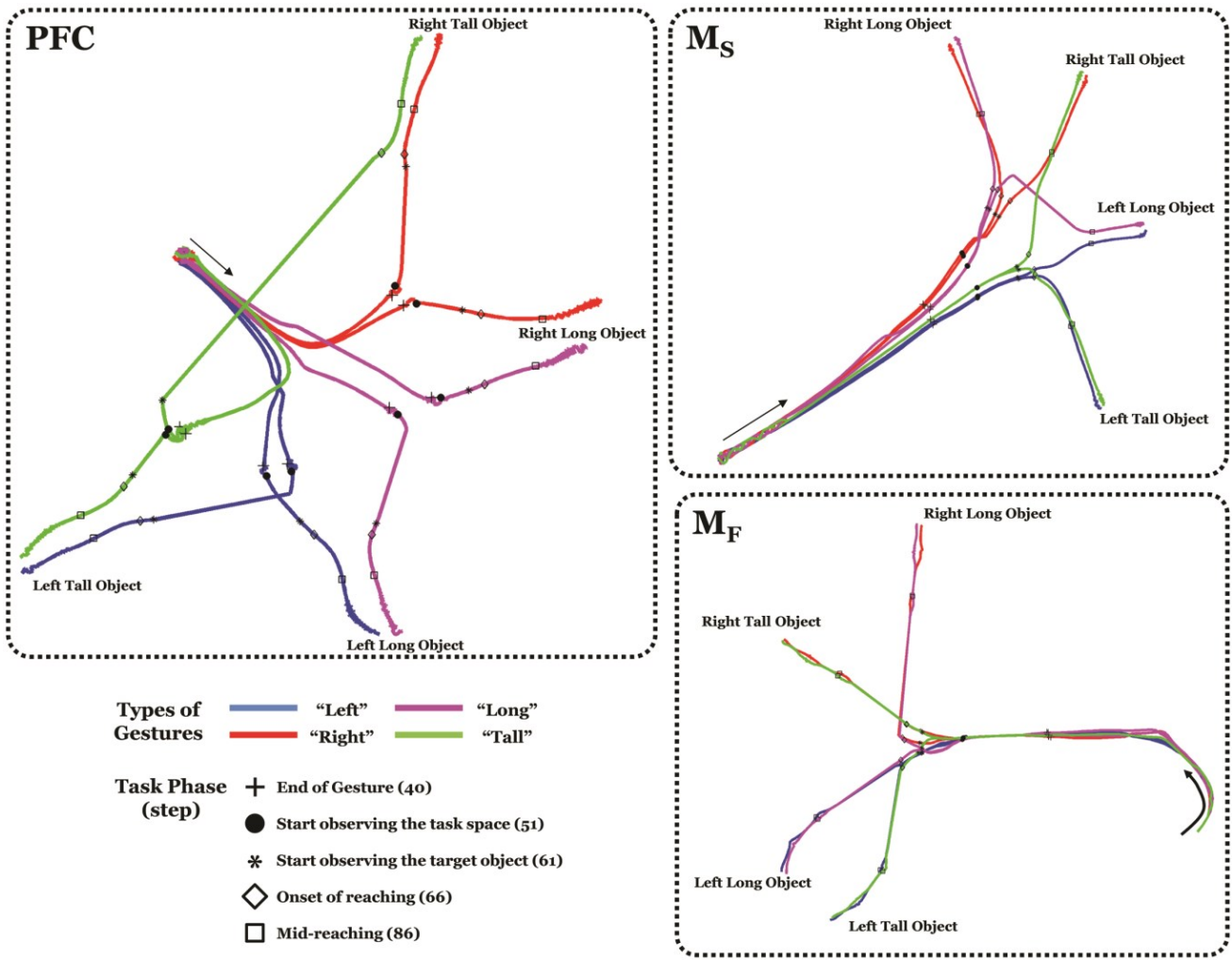

Fig. 5. The development of internal representations in PFC, $M_S$ and $M_F$ for exemplar cases. 8 trials consisting of 4 different configurations of the target object indicated by 2 different types of human gesture were examined using the t-SNE algorithm. Line colors indicate the type of human gesture, markers indicate the task phase and an arrow indicates the direction of the time step. At the end of the trajectories, the configuration of each target object is specified.

We further analyzed the development of internal representations for exemplar cases in the PFC, $M_S$ and $M_F$ layers using the t-SNE algorithm (Fig. 5). A total number of 8 representative training trials consisting of 4 different target objects indicated by 2 different human gestures were compared. In the t-SNE analysis, the number of the initial dimensions and the perplexity were set to 10 and 50 respectively. Fig. 5 illustrates that the proposed model dynamically integrated perception and computed motor plans. The PFC layer developed higher-level task-related representations encompassing both visual and motor information whenever available. For instance, the PFC layer identified the type of gesture even before the end of human gesture presentation. While the robot was observing the gestures, the representations at the PFC layer already started to develop differently according to the type of gesture. When the robot started observing the task space, representations developed differently depending on the object's features, such as location and type. Analysis shows that the $M_S$ layer was calibrated based on the perceived human gesture. For example, when the robot started observing the task space, representations in the $M_S$ layer began to differentiate themselves depending on the perceived human gesture. The implication here is that higher-level proprioception was calibrated based on perceived visual information. The representations of the $M_F$ layer developed similarly to those of the $M_S$ layer, showing very similar development of representations for the same target object, but they were more closely related the robot's current action. In sum, the $M_S$ layer mainly encoded higher-level motor actions and mediated between higher-level cognition (PFC) and the lower-level motor actions ($M_F$).

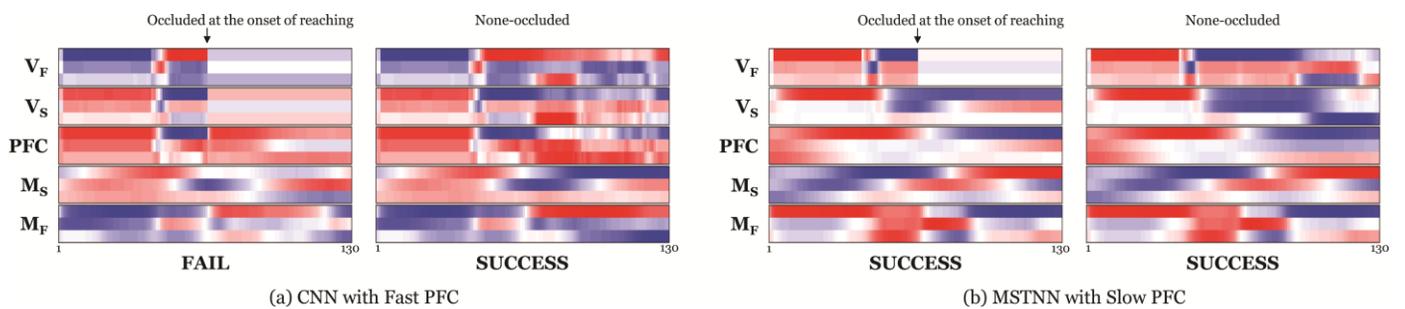

Fig. 6. The development of the first three principal components (PCs) at each layer in the two opposite cases: (a) the CNN with the fast PFC condition and (b) the MSTNN with the slow PFC condition. The numbers at the bottom (horizontal lines) indicate the time step from the beginning of the task to the end of the task, and the colors denote component score values. Legends are omitted since as our focus is on similarities in the development of neuronal activation with different occlusion timings.

In order to clarify differences in the internal dynamics of the model when vision input was occluded, we compared two different timings in the two diametrically opposed cases: not occluded and occluded at the onset of reaching, in the CNN with the fast PFC condition and in the MSTNN with the slow PFC condition. The development of the first three principal components (PCs) of each layer for the same target object is depicted from the beginning to the end of the task in Fig. 6. In the CNN with fast PFC condition, the robot failed to grasp the target object when visual input was completely occluded at the onset of reaching. However, it was able to grasp the target object when the visual input was not occluded. Analysis clearly reveals that PCs in the PFC, $M_S$ and $M_F$ layers developed differently in both cases, implying that the model cannot form consistent representations. In the MSTNN with the slow PFC condition, the robot was able to grasp the target object in both cases. Analysis reveals that the development of PCs of the PFC, $M_S$ and $M_F$ layers in occluded and non-occluded cases were similar. Especially, when vision input was occluded at the onset of reaching, the activation of $M_S$ and $M_F$ developed similarly with those in the non-occluded case, implying the development of the coherent proprioceptive representations regardless of occlusion conditions. In sum, the PFC layer as well as MTRNN layers developed consistent representations even when visual information was lost at the initiation of a reaching action, resulting in successful task performance.

## VI. Discussion

Throughout the experiments, we verified several key aspects of the proposed model. In this section, we discuss them in detail.

### A. Self-organized Coordinated Dynamic Structure

The proposed model developed a coordinated dynamic mechanism in the whole network, enabling the robot to learn goal-directed behaviors through the seamless coordination of cognitive skills. In terms of downward causation, spatio-temporal constraints imposed on the each level of the hierarchy and end-to-end learning performed on the tightly coupled structure are the two key factors in achieving the coordinated dynamic mechanism.

First, multiple scales of spatio-temporal constraint enabled the model to dynamically compose conceptual representations in the higher level by assembling perceptual features in the lower level, suggesting the emergence of different cognitive functionalities in its hierarchy. Sporns [48] argues that cognitive functions develop in human brains through anatomical spatio-temporal constraints including connectivity and timescales among local regions. Similarly, different spatio-temporal constraints imposed on different parts of the proposed model lead to the development of different cognitive functionalities at each level of the hierarchy. For example, the PFC layer mainly encoded higher-level task-related information whereas the MTRNN layers encoded information related to the robot's current action. Specifically, lower-level proprioception layer ($M_F$) developed the representation closely reflecting the robot's current action whereas higher-level proprioception layer ($M_S$) played the role of mediating between higher-level cognition (PFC) and lower-level proprioception ($M_F$). This result is analogous to findings in [49] that higher-level task-related information was encoded in the PFC and the lower-level arm movements were encoded in the primary motor cortex of monkeys.

Second, end-to-end learning performed on a tightly coupled structure enabled the model to form coordinated representations without explicitly encapsulating each perceptual modality, i.e. perception, action, and decision making. On the present model, the coupling of perception and action is achieved by connecting two different deep networks through the PFC layer. This coupling enabled the model to generate motor plans dynamically in response to visual perceptual flow. This finding is in line with the previous studies [13, 40] in which the importance of coupling perception and action was emphasized. Analysis of neural activation shows that this dynamic motor plan generation involved continuous integration of the visual perceptual flow, rather than being directly mapped from visual perception. This dynamic transformation of visual information into behavioral information was also observed in the experiments with macaque monkey's brains [50].

Multimodal representation helps with distinguishing object type and orientation [3] and competency for multimodal information integration is considered essential for embodied cognition [2, 4, 51, 52]. Ultimately, the present model was able to develop multimodal representations by abstracting and associating visual perception with proprioceptive information in different pathways. In other words, higher level representations were based not only on visual information about and object and corresponding human gesture, but also on information about movement trajectories resulting in the successful grasp of the target object.

*B. Memory and Pre-planning Capability*

Results indicate that the proposed model is capable of developing and employing working memory. We found that the robot was able to maintain task-related information in higher levels throughout the task phases, and dynamically combining it with object percepts. For instance, the robot maintained human intention categorized at the beginning of the task and combined it with the object percept, so that it could reach and grasp the target object. Furthermore, the proposed model performed robustly even when visual input was completely and unexpectedly occluded. This memory capability was achieved by the temporal hierarchy of the model as well as by the recurrent connections of the PFC layer. Particularly, when the time constants of the higher-level were larger than those of the lower-level, the model showed the most robust performance in the various circumstances including the experiments with the novel object configurations as well as the ones with unexpected visual occlusion. Although the suitable values of time constant at each level might differ depending on the task, the finding of our study suggests that the progressively larger time constants from the low-level to the higher-level of the architecture played an important role to form the coordinated dynamic structure including a memory capability. This finding is also in consistency with the other previous studies [9, 17-18, 33-34, 36-37, 46] that have shown the importance of a similar temporal hierarchy in the multiple timescales neural network such as MSTNN and MTRNN. This internal contextual dynamics of the proposed model highlights a key difference between the proposed model and the previous study [13] which lacked capability for keeping memory.

Furthermore, the model was able to pre-plan an action prior to action execution. More specifically, neuronal activation in the higher-level proprioception layer ($M_S$) was calibrated based on the perceived visual images prior to action execution. This pre-planning capability played a particularly important role in the visual occlusion experiment, enabling the robot to reach and grasp the target object even without monitoring the target object and the hand during reaching. This result is analogous to findings in [53] which reported that F5 neurons in the brain of the macaque monkey encoded grip-specific information even when no movement was intended.

## VII. Conclusion

The current study introduced the Visuo-Motor Deep Dynamic Neural Network (VMDNN) model which can learn to read human intention and generate corresponding behaviors in robots by coordinating multiple cognitive processes including visual recognition, attention switching, memorizing and retrieving with working memory, action preparation and generation in a seamless manner. The simulation study on the model using the iCub simulator revealed that the robot could categorize human intention through observing gestures, preserve the contextual information, and execute corresponding goal-directed actions. The analysis further showed that synergic coordination among these cognitive processes can be developed when end-to-end learning of tutored experience is performed on the whole network, allowing dense interaction between subnetworks. In conclusion, the aforementioned cognitive mechanism can be developed by means of downward causation in terms of spatio-temporal scale differentiation among local subnetworks, topological connectivity among them, and the way of interacting through sensory-motor coupling.

There are several research directions suggested by this study. First, experiments incorporating other cognitive skills as well as other sensory modalities need to be conducted for a better understanding of the mechanisms of learning goal-directed actions in human and other biological neural systems. Second, the scalability of the proposed model will need to be examined in experiments with a real robot and with a larger variety of objects. The complexity of the task that the model can perform seems proportional to the size of the learnable parameters including the number of layers in each subnetwork and the number of neurons in each layer. Third, the robot's sensorimotor experience in our study was acquired from tutoring by the experimenter. In some cases, tutoring may be impossible, so it would be worth investigating how such experience can be obtained autonomously.